\documentclass[11pt,letterpaper]{mystyle}
\usepackage[comma,authoryear,compress]{natbib}
\usepackage{array,multirow}
\hypersetup{citecolor=YaleBlue,pdftitle={HabitusMem: Belief Memory over Evolving User Preferences for LLM Agents}}

\newcommand{\passall}{\mathrm{Pass}^{4}}
\newcommand{\passany}{\mathrm{Pass}@4}
\newcommand{\avgfour}{\mathrm{Avg}@4}
\newcommand{\code}[1]{\texttt{#1}}
\definecolor{qabmteal}{HTML}{166A73}
\definecolor{qabmgray}{HTML}{6D7885}
\definecolor{qabmgold}{HTML}{B77926}
\definecolor{qabmblue}{HTML}{E6F0FA}
\definecolor{qabmred}{HTML}{B5473C}
\newtcolorbox{casecard}[2]{colback=white,colframe=qabmteal,colbacktitle=qabmteal,coltitle=white,fonttitle=\small\bfseries,boxrule=.5pt,arc=2pt,left=6pt,right=6pt,top=5pt,bottom=3pt,before skip=9pt,after skip=9pt,title={#1\hfill{\mdseries\footnotesize #2}}}
\newcommand{\casefield}[2]{\noindent\parbox[t]{1.6cm}{\raggedright\footnotesize\bfseries\textcolor{qabmteal}{#1}}\hfill\parbox[t]{\dimexpr\linewidth-1.8cm\relax}{\small #2}\par\vspace{3.5pt}}
\newcommand{\beliefrecord}[4]{{\raggedright\ttfamily\footnotesize\frenchspacing\hangindent=1em\hangafter=1 - #1 \textcolor{black!62}{| evidence: #2 | scope: #3 | status:} #4\par\vspace{3pt}}}
\newcommand{\statusactive}{\textcolor{qabmteal}{active}}
\newcommand{\statusretracted}{\textcolor{qabmred}{retracted}}

\title{HabitusMem: Belief Memory over Evolving User Preferences for LLM Agents}
\runningtitle{HabitusMem: Belief Memory over Evolving User Preferences for LLM Agents}
\author{Yifan Lu\textsuperscript{1,*}, Gaoyang Guo\textsuperscript{2}, Wen Chen\textsuperscript{2,\textdagger},
    Ziyun Zou\textsuperscript{1}, Jian Wu\textsuperscript{2},
    Han Zhu\textsuperscript{2}, Imran Razzak\textsuperscript{1}\\
    {\Affilfont
    \textsuperscript{1}Mohamed bin Zayed University of Artificial Intelligence\\
    \textsuperscript{2}Alibaba Group}}
\primarycontact{yifan.lu@mbzuai.ac.ae}
\correspondingauthor{Wen Chen (\href{mailto:chenyu.cw@alibaba-inc.com}{chenyu.cw@alibaba-inc.com})}
\hypersetup{pdfauthor={Yifan Lu, Gaoyang Guo, Wen Chen, Ziyun Zou, Jian Wu, Han Zhu, Imran Razzak}}
\coderepository{https://github.com/csyifan/HabitusMem}

\begin{document}

\begin{abstract}
Personalized LLM agents must act on a user's current preferences, which change across interactions and are rarely stated in full. Existing memory systems store and retrieve content from the interaction history and leave the agent to work out which preferences still hold and where they apply. Such content does not indicate whether a preference is still current or where it applies, so agents often act on outdated or misapplied preferences. We instead treat memory as the assistant's belief about the user. Following Bourdieu's account of habitus as durable but revisable dispositions, we model the user as a latent preference state that every interaction, including the current request, both reveals and updates. The belief summarizes what the history supports about this state, and memory consists of the individual beliefs that bear on the agent's current decision, each with its evidence, scope, and status. We implement this view in HabitusMem, a trained writer that states these beliefs as text records for a frozen agent. Across two agent families, HabitusMem achieves the best results among memory systems on both PersonaMem-v2 and VitaBench 2.0. Controlled analyses show that learned request conditioning drives the answering gain, while auxiliary training tasks improve tool use.
\end{abstract}

\maketitle
\section{Introduction}
\label{sec:intro}
Large language model (LLM) agents increasingly assist the user across many interactions, answering questions and carrying out tasks on the user's behalf \citep{personamemv1,personamemv2,vitabench2}. Useful assistance depends on preferences and constraints that a new request rarely states in full. These preferences form through repeated choices, persist across situations, and change over time: a user who has moved supplies a new delivery address, a user who has lost interest in a topic asks the assistant to stop raising it, and a gift chosen for a friend reveals little about the user's own taste. The agent must therefore act on the user's current preferences, which it can only infer from what the user has said and done.

\begin{figure}[t]
\centering
\includegraphics[width=\linewidth]{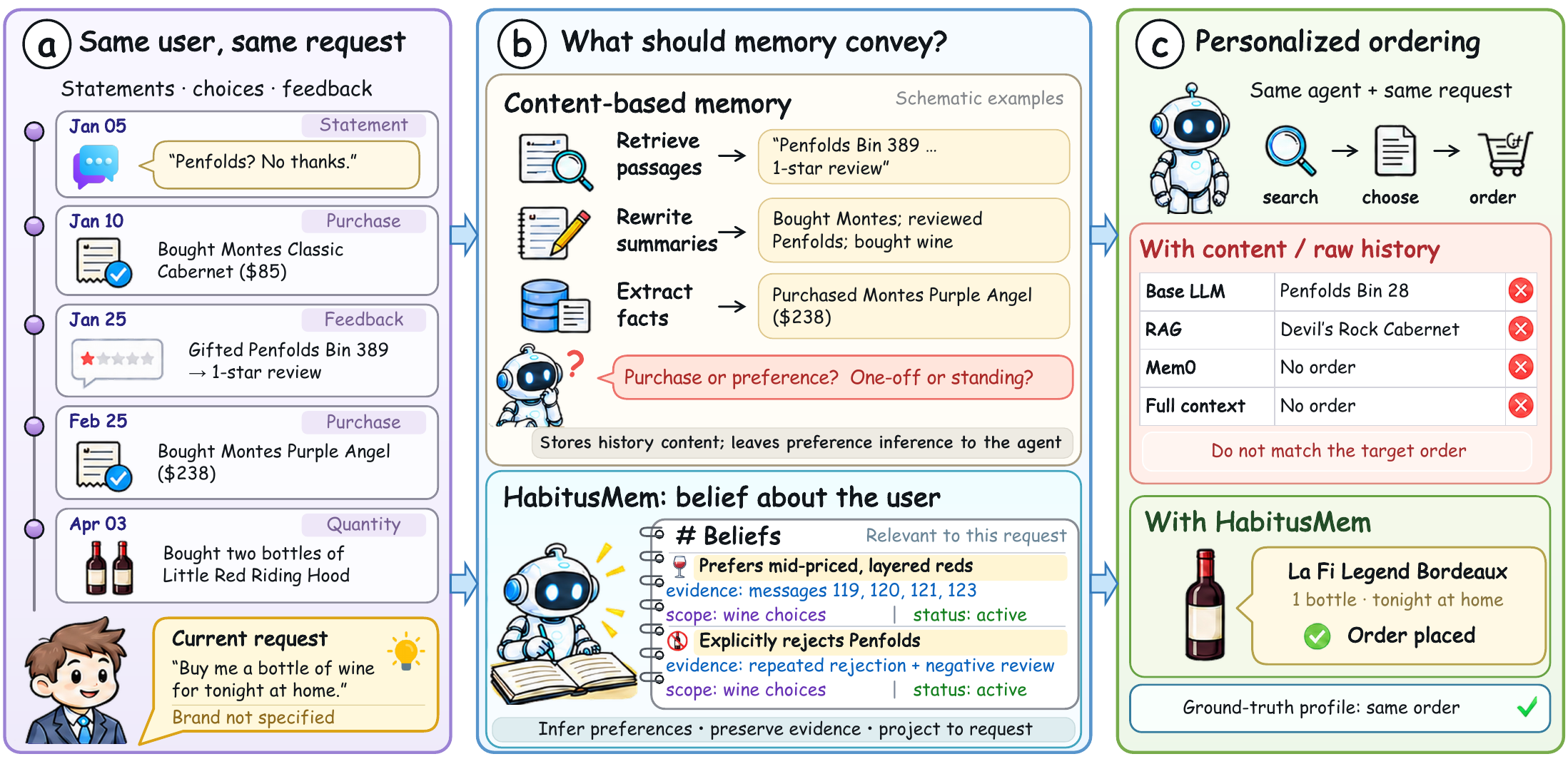}
\caption{\textbf{Content-based memory method versus belief memory on a personalized order.} The user has repeatedly rejected Penfolds wine and now asks for a bottle of wine without naming a brand. Content-based memory methods pass on fragments such as past purchases and a negative review, and agents using them order the wrong wine or none at all. HabitusMem infers that the user prefers mid-priced layered reds and rejects Penfolds, records the evidence, scope, and status of each belief, and leads the agent to the same order as the ground-truth profile.}
\label{fig:case}
\end{figure}

Existing memory systems carry this information across interactions by storing content derived from the history: they extract facts, maintain notes or summaries, or retrieve and compress past passages for the current request \citep{memorybank,mem0,amem,rag,longllmlingua}. Such memory records what the user said or did rather than what the history implies about the user now (Figure~\ref{fig:case}), and it leaves three questions to the agent. \emph{How strongly does the history support a claim?} Memory stores claims without their evidential basis, so a disposition confirmed across many interactions and an incidental or indirect observation enter in the same form, and the agent cannot weigh one against the other. \emph{Where does a claim apply, and does it still hold?} Entries carry no scope or status. Update operations overwrite or delete earlier entries and thus erase the difference between a preference that has changed and one that has only been withdrawn; without them, contradictory entries coexist with nothing to mark which is current, and a context-specific instruction cannot be told apart from a standing preference. \emph{Which claims matter for the decision at hand?} Memory is written or retrieved by salience and surface similarity rather than by relevance to the agent's decision, so it can drop constraints the decision depends on and abstract away details that the action requires exactly.

The underlying problem is that these designs represent the history, whereas personalization needs a representation of the user. Bourdieu's concept of \emph{habitus} describes what personalization targets: durable dispositions, formed through past experience, that generate practices in concrete situations \citep{bourdieu1977outline}. These dispositions are durable but revisable, since new experiences are perceived through categories formed by earlier ones and in turn reinforce or modify them \citep[p.~133]{bourdieu1992invitation}. For an assistant, this suggests modeling the user as a latent preference state shaped by the whole history and carried forward by every interaction, including the current request. A belief-state view makes this concrete \citep{berger1985statistical}: earlier interactions form the assistant's belief about the user, each new interaction revises it, and the current belief summarizes what the history supports about the user's present state. We express this belief as a set of individual beliefs, each a claim about the user's preferences, circumstances, or constraints, and define memory as the subset of these beliefs that bears on the agent's current decision, since the agent acts to minimize expected loss under the belief. Memory is thus composed of beliefs about the user rather than of content from the history. Each of the three questions then has a direct answer. A claim's \emph{evidence} conveys its credibility; its \emph{scope} and \emph{status} convey where it applies and whether later observations have left it active, superseded, or retracted; and the decision posed by the request determines which claims to include.

We instantiate this view in HabitusMem, a writer that composes memory from belief records of the form \code{claim | evidence | scope | status}, one belief per record, for a frozen agent. Reading the raw history and the current request, the writer answers the three questions directly: it cites the observations that support each claim, marks its scope and whether later interactions have superseded or retracted it instead of overwriting it, and keeps only the claims that bear on the request, with the exact choices and constraints an action requires. A single model learns these operations from four tasks that mirror belief tracking: direct writing infers claims and their evidence, state update revises them with new interactions, and request and action projection select those a question or transaction depends on. The resulting memory works with different agents without modifying them. Across two agent families, HabitusMem achieves the best results among memory systems on PersonaMem-v2 and VitaBench 2.0, and controlled analyses attribute its gains to learned request conditioning and the auxiliary tasks. Our contributions are as follows:
\begin{itemize}
\item We reframe personalized memory as the assistant’s belief about the user following Bourdieu’s account of habitus. The resulting memory states how credible each preference is, where it applies, whether it has changed or been withdrawn, and whether it matters for the current decision, enabling the agent to act on the user's current preferences.
\item We introduce HabitusMem, a writer that states each belief as a text record with its evidence, scope, and status for a frozen agent. A single adapter learns belief inference, belief update, and decision projection from four training tasks, which supports both personalized answering and task execution that depends on exact constraints.
\item On PersonaMem-v2 and VitaBench 2.0, with two agent families, HabitusMem is the most accurate compact memory for personalized answering and the most consistent for personalized tool use, where it also surpasses giving the agent the full history.
\end{itemize}

\section{Related Work}
\paragraph{Memory for LLM agents.}
Agent memory systems differ in their substrate, their cognitive analogy, and whose experience they store \citep{huang2026agentmemory}. User-centric systems maintain external stores of extracted facts, notes, or summaries that are retrieved for later requests \citep{memorybank,memgpt,mem0,amem,zep}. Other designs organize memory after human cognition, for example through episodic event segmentation or Hebbian association \citep{ICLR2025_c05144b6,zhu-etal-2026-hela}, and agent-centric memory accumulates the agent's own task experience \citep{voyager,reasoningbank,awm}. A growing line learns memory operations, either by writing knowledge into model parameters \citep{memoryllm,wise} or by optimizing memory edits against downstream reward with reinforcement learning \citep{memoryr1,memalpha,memagent}. Retrieval augmentation and query-aware compression instead select or shorten passages for the current input \citep{rag,longllmlingua,recomp}. Across these designs, an entry records what was said or observed without stating how credible a claim is, where it applies, or whether a later observation has superseded or retracted it. HabitusMem keeps the raw history as the source of evidence and defines memory as the part of the assistant's belief about the user that bears on the current decision, with these properties stated in each record.

\paragraph{Personalized agents and benchmarks.}
Personalized agents extend LLMs with user profiles, memory, planning, and action execution \citep{xu2026personalizedagents}. Profile modeling builds persona descriptions of stable traits \citep{fspo,rlpa,curio} or factorizes preferences over candidate responses \citep{rfm,pref,lore}, and agent designs adapt to the user or act on the user's behalf \citep{ldagent,personaagent,personax}. CAPE learns population-level preference priors and uses Bayesian inference to guide cold-start preference elicitation \citep{bose2026pep}. \citet{westhaeusser2025persistentprofiles} combine persistent memory with continuously updated user profiles, query-dependent retrieval, and self-validation. Personalized planning conditions decisions on profiles and memory in a single pass \citep{prime,alignxplore,pbp} or refines plans through clarification, user revision, or environment feedback \citep{sageagent,askbeforeplan,aipom}, and personalized execution grounds tool selection and arguments in user preferences \citep{petoolllm,toolspectrum,awareus} or re-ranks and proactively suggests outcomes \citep{pear,etapp}. These components consume user memory but leave its content to summarization or similarity-based retrieval, which does not separate credible, applicable, and current preferences from the rest. Benchmarks have moved from profile-conditioned generation \citep{lamp} and long-term conversational recall \citep{locomo,longmemeval} to preferences that are implicit and evolving \citep{personamemv1,personamemv2} and must be applied in tool-using tasks \citep{vitabench2}. We evaluate on the last two settings, in which the agent must infer which preferences currently hold before acting on them.

\section{Methodology}
\label{sec:method}
\begin{figure}[t]
\centering
\includegraphics[width=\linewidth]{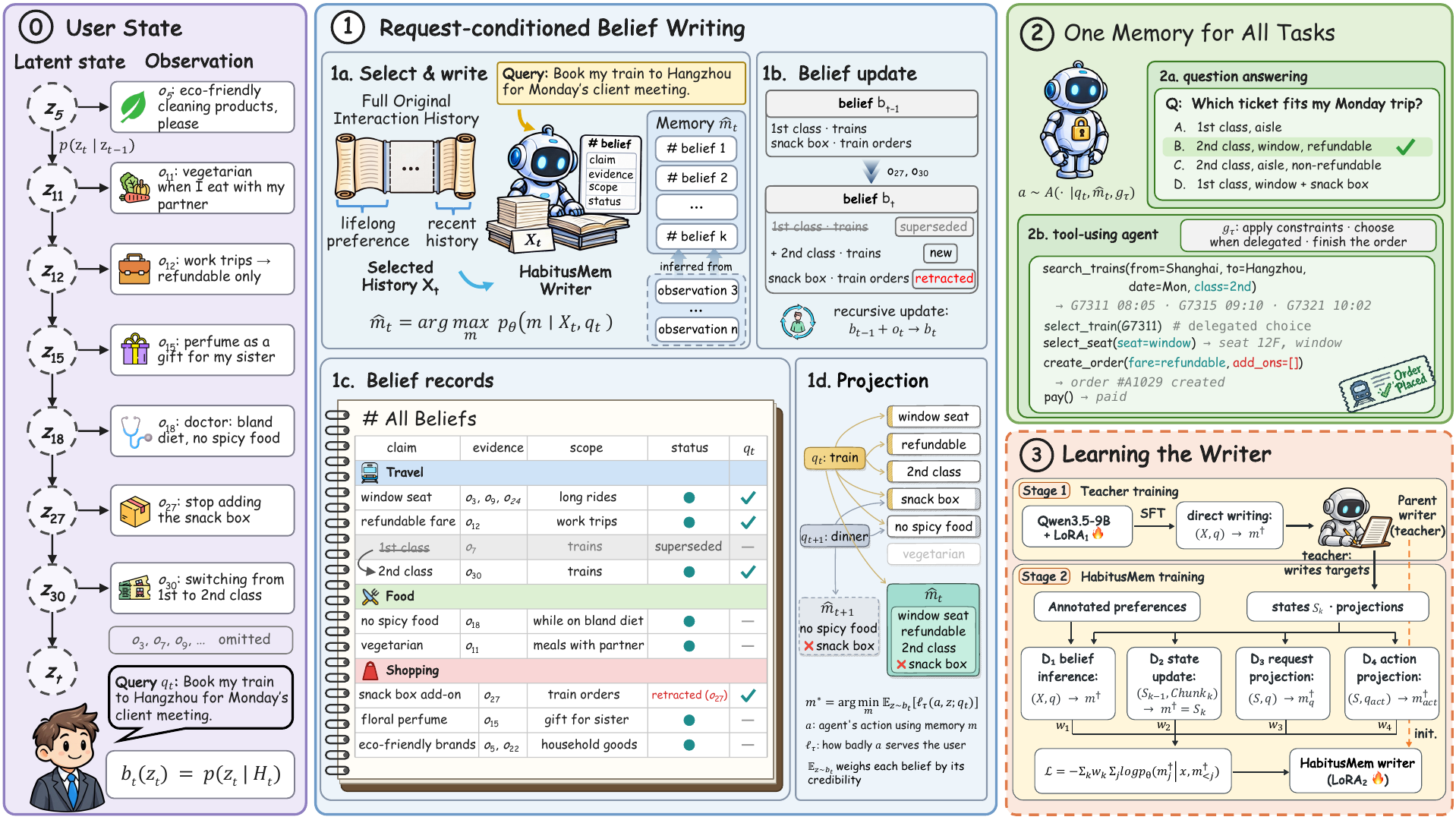}
\caption{\textbf{Overview of HabitusMem.} The writer takes the retained raw interactions and the current user query as input and infers belief records, each with its evidence, scope, and status. It then projects only the beliefs that bear on the user query to the agent, which answers or acts on the user query. Here, the bland diet and the late-shift hospital address shape the dinner order, whereas the partner-specific vegetarian preference and the one-off oat-milk instruction are kept but not projected. A single writer learns these steps from four tasks: direct writing, state update, request projection, and action projection.}
\label{fig:overview}
\end{figure}
HabitusMem composes memory from the assistant's beliefs about the user's current preference state, selected for the decision the agent faces. As Figure~\ref{fig:overview} shows, a single writer reads the raw interactions together with the current request, infers beliefs with their evidence, scope, and status, and passes those that bear on the request to a frozen agent. Section~\ref{sec:formulation} formalizes the belief and the decision it serves, Section~\ref{sec:construction} describes how the writer expresses the belief as text records, and Section~\ref{sec:training} describes how it is trained to infer, revise, and select beliefs.

\subsection{Problem formulation}
\label{sec:formulation}
Let $z_t$ denote the user's preference state at interaction $t$, comprising the durable dispositions, current circumstances, and explicit constraints that shape what the user wants. The state evolves: a new experience may reinforce a disposition, modify it, or lead the user to withdraw it, which we describe by a transition $p(z_t\mid z_{t-1})$. The assistant observes the state only through interactions $o_t\sim p(o_t\mid z_t)$, such as statements, choices, corrections, and requests. The current request $q_t$ is the most recent of these observations: it carries evidence about the user and also poses the decision that the agent must make. Given the history $H_t=(o_1,\ldots,o_t)$, the belief that the history supports is
\begin{equation}
b_t(z_t)=p(z_t\mid H_t)\;\propto\;p(o_t\mid z_t)\int p(z_t\mid z_{t-1})\,b_{t-1}(z_{t-1})\,dz_{t-1}.
\label{eq:posterior}
\end{equation}
The recursion mirrors the account of habitus, in which each observation is interpreted through the belief formed by earlier ones and either reinforces or revises that belief. The same belief can equally be read from the complete history at once, and we use Equation~\ref{eq:posterior} as a conceptual account of how the belief should respond to evidence.

The belief describes the user, and memory conveys it to the agent that acts. Given the request $q_t$, a memory $m$, and optional execution instructions $g_\tau$ for task type $\tau$, a frozen agent produces an answer or action trajectory $a\sim A(\cdot\mid q_t,m,g_\tau)$, and a loss $\ell_\tau(a,z_t;q_t)$ measures how poorly $a$ serves a user in state $z_t$. We score a memory by the agent's expected loss under the belief and take the ideal memory to be the one that minimizes it,
\begin{equation}
\mathcal R_\tau(m;H_t)
=\mathbb E_{z_t\sim b_t}\,
\mathbb E_{a\sim A(\cdot\mid q_t,m,g_\tau)}
\left[\ell_\tau(a,z_t;q_t)\right],
\qquad
m^\star_t\in\arg\min_{m\in\mathcal M}\mathcal R_\tau(m;H_t),
\label{eq:objective}
\end{equation}
where $\mathcal M$ is the set of textual memories. Equation~\ref{eq:objective} separates two questions that content-based memory conflates. The belief determines how credible each claim about the user is, and the request determines, through the loss, which claims change the agent's decision. A gardening preference is equally credible whether the user asks for a weekend activity or a train ticket, but only the first request makes it relevant. Conversely, an order can fail on a single wrong variant or quantity, so the memory must convey such details exactly. Under a fixed message budget, conditioning the memory on the request can only lower the best attainable risk.

The formulation also determines what a textual memory should state. Each record in HabitusMem has four fields. The \emph{claim} states a fact, preference, or constraint about the user. Its \emph{evidence}, the observations that support it, lets the agent distinguish a disposition established by repeated explicit statements from one inferred from a single choice. Its \emph{scope}, the situations in which the disposition guides behavior, prevents a choice made for one occasion from steering every later request, since a disposition is expressed differently across situations. Its \emph{status} records how the state has evolved, marking whether the claim remains active, has been superseded by a later preference, or has been retracted because the user asked the assistant to stop using it, which restricts its use without making it false.

Equations~\ref{eq:posterior} and~\ref{eq:objective} serve as a design rationale that specifies what the writer should produce. Exact inference over natural-language histories is intractable, and HabitusMem neither estimates numerical posteriors nor optimizes Equation~\ref{eq:objective} through agent rollouts. Instead, a language-model writer is trained on reference memories to map a history and a request directly to belief records, and Appendix~\ref{app:bayesian} discusses how these records relate to the formulation.

\subsection{Memory construction}
\label{sec:construction}
Memory construction starts from the accumulated raw interaction log. New observations are appended to the log, and a position- and budget-based rule $T_\tau$ selects the view $X_t=T_\tau(H_t)$ that fits the writer's input: on PersonaMem-v2 it keeps the beginning of each history, and on VitaBench 2.0 it keeps the beginning together with the most recent interactions. This rule determines which evidence the writer can see, and the writer decides what enters the memory.

The writer receives the selected history, a format instruction $I$, and a suffix $s_\tau(q_t)$ containing the current request. For question answering, the suffix asks for facts and preferences that help answer the request. For agent tasks, an action suffix asks for explicit choices and fulfillment constraints, such as products, variants, quantities, addresses, and times, and asks the writer to keep explicit historical choices rather than inferred alternatives.

The output is a list of records under a \code{\#\# Beliefs} header, one per line in the form \code{claim | evidence: source | scope: context | status: state}. The four fields are those defined above. Together, the fields separate an enduring disposition, a one-off instruction, and a withdrawn preference. For example, an oat-milk order scoped to \emph{this delivery only} keeps its exact quantity without steering later requests, and a spicy-food preference marked \emph{retracted (message 61)} no longer guides restaurant choices.

We instantiate the writer with Qwen3.5-9B and one LoRA adapter shared across tasks. The writer generates the memory in one greedy decoding pass, and the agent acts on the records together with the current request,
\begin{equation}
\hat m_t=\operatorname{GreedyDecode}\big[p_\theta(\cdot\mid I,X_t,s_\tau(q_t))\big],
\qquad
\hat a\sim A(\cdot\mid q_t,\hat m_t,g_\tau).
\label{eq:inference}
\end{equation}
The execution instructions $g_\tau$ direct the agent to apply relevant constraints, to make a choice when the user delegates one, and to complete the requested transaction. Whenever the history or the request changes, the writer rewrites the memory from the selected raw log and keeps no persistent state between requests; repeated reads with the same history and request reuse the cached memory.

\subsection{Training}
\label{sec:training}
We train the writer in two stages on the official PersonaMem-v2 training split. The first stage produces a parent writer that infers beliefs from a history and a request. The second stage uses this parent as a frozen teacher to construct examples of the operations it lacks, belief update and decision projection, and trains HabitusMem on all of these operations together.

\paragraph{Stage 1: Parent writer training.}
Writing a useful belief requires both the record format and the ability to identify the preference a request depends on, so we teach them in turn. We first cold-start a Qwen3.5-9B LoRA adapter on belief memories from a frozen DeepSeek-V4-Flash teacher, which writes one schema-valid memory at each of two positions in every training history. This substep teaches the record syntax without using preference annotations. The resulting writer then generates a static belief memory for each persona, and for each training request we place the request's annotated preference at the top of this memory as a new record. The status of the record follows the annotation, marking a preference as active, as superseding an earlier one, or as retracted, in which case the record names the topic but not the revoked content. Training on these memories teaches the parent to surface the belief that a request depends on. The parent sees no update or projection examples, so it cannot revise stored records or select them for a decision.

\paragraph{Stage 2: HabitusMem writer training.}
We initialize HabitusMem from the parent and train it on four tasks that cover inferring, revising, and selecting beliefs. \emph{Direct writing} $\mathcal D_1$ retains the Stage 1 targets and teaches belief inference from evidence. \emph{State update} $\mathcal D_2$ teaches belief revision. The parent processes each history chunk by chunk, a data builder merges the resulting records into a running state $S_k$, and each transition $(S_{k-1},\text{chunk}_k)\to S_k$ yields an example of superseding or retracting earlier claims as new evidence arrives. \emph{Request projection} $\mathcal D_3$ and \emph{action projection} $\mathcal D_4$ teach the selection of beliefs for the current decision. From the final state, the parent writes the view that a request needs, and for requests that require execution the view must also preserve the exact choices and fulfillment constraints. Target preparation enforces the record syntax but does not check the factual content of the records. The writer minimizes the teacher-forced token loss
\begin{equation}
\mathcal L(\theta)
=-\sum_{k=1}^{4}w_k\sum_{(x,m^\dagger)\in\mathcal D_k}\sum_j
\log p_\theta(m_j^\dagger\mid x,m^\dagger_{<j}),
\label{eq:sft}
\end{equation}
where $w_k$ is the weight of task $k$. All four tasks update one adapter, so that inference, update, and projection share parameters. Appendix~\ref{app:training} gives the target construction and training settings, and Appendix~\ref{app:mixture-sensitivity} compares task weights.

\section{Experiments}
\label{sec:experiments}
We evaluate HabitusMem on PersonaMem-v2 for personalized question answering and on VitaBench 2.0 for personalized tool use. Section~\ref{sec:main-results} compares it with established memory systems, and Section~\ref{sec:components} isolates the contributions of request conditioning, the auxiliary tasks, and the record fields. Appendices~\ref{app:sensitivity}--\ref{app:cases} further report training-mixture and inference sensitivity, computational cost, controlled probes of evidence, scope, and status, and case studies.

\subsection{Experimental setup}
\label{sec:setup}
\paragraph{Benchmarks and metrics.}
PersonaMem-v2 \citep{personamemv2} evaluates personalized answering from long interaction histories, including changing preferences and requests to forget. We evaluate in the English 32k setting, whose test split contains 5,000 queries. The test queries span nine task scenarios. We report personalized QA accuracy on multiple-choice questions as the primary metric and a generation score assigned by an external Qwen3.5-9B judge. VitaBench 2.0 \citep{vitabench2} contains 771 subtasks from 56 users, in which an agent uses tools to complete delivery, in-store consumption, and online travel tasks according to preferences implicit in past dialogues and behavior logs. We run four trials per subtask, with an external Qwen3.5-9B simulating the user and scoring the trajectory against the subtask rubric. We report average reward ($\avgfour$), success in at least one of four trials ($\passany$, \emph{any-success}), and success in all four ($\passall$, \emph{all-success}).

\paragraph{Agents and comparison systems.}
We test personalization with two locally deployed agents, Qwen3.5-397B-A17B and DeepSeek-V4-Flash-0731. We compare with five compact memory systems: retrieval-augmented memory (RAG) \citep{rag}, Rewrite (an LLM-rewritten history summary), Mem0 \citep{mem0}, A-MEM \citep{amem}, and MemListener \citep{memlistener}. The base LLM receives no memory, and full context gives the agent direct access to the interaction history. On VitaBench 2.0, a ground-truth profile built from annotated preferences is an additional privileged reference. RAG and Rewrite are implemented in the evaluation harness.

\subsection{Main results}
\label{sec:main-results}
\begin{table*}[!t]
\centering\footnotesize
\setlength{\tabcolsep}{3.4pt}
\caption{Personalized QA accuracy on PersonaMem-v2 by task scenario. Bold and underlined values mark the best and second-best compact memory systems in each column.}
\label{tab:pm-baselines}
\begin{tabular}{@{}lcccccccccc@{}}
\toprule
\multirow{2}{*}{Memory} & Know- & Pro- & Pro- & Crea- & Trans- & Personal- & \multirow{2}{*}{Trouble} & \multirow{2}{*}{Chat} & \multirow{2}{*}{Social} & \multirow{2}{*}{Overall}\\
& ledge & Email & Writing & tive & lation & Email & & & & \\
\midrule
\multicolumn{11}{@{}l}{\textit{Qwen3.5-397B-A17B}}\\
Base LLM & 0.644 & 0.408 & 0.381 & 0.420 & 0.438 & 0.433 & 0.393 & 0.396 & 0.386 & 0.446\\
Full context & 0.519 & 0.594 & 0.573 & 0.543 & 0.624 & 0.585 & 0.599 & 0.548 & 0.565 & 0.569\\
\cmidrule(l){1-11}
RAG & 0.436 & 0.481 & 0.453 & 0.462 & 0.486 & 0.470 & 0.437 & 0.450 & 0.474 & 0.460\\
Rewrite & \underline{0.519} & \underline{0.503} & 0.466 & 0.466 & 0.479 & 0.469 & \textbf{0.509} & 0.440 & 0.485 & 0.484\\
Mem0 & 0.490 & 0.449 & \underline{0.488} & 0.471 & \underline{0.533} & 0.453 & 0.475 & 0.438 & 0.415 & 0.470\\
A-MEM & 0.433 & 0.494 & 0.457 & \textbf{0.510} & 0.503 & \textbf{0.530} & \underline{0.497} & \textbf{0.502} & \underline{0.515} & \underline{0.489}\\
MemListener & 0.443 & 0.399 & 0.422 & 0.373 & 0.438 & 0.404 & 0.459 & 0.388 & 0.423 & 0.418\\
\rowcolor{qabmblue}HabitusMem & \textbf{0.649} & \textbf{0.526} & \textbf{0.529} & \underline{0.503} & \textbf{0.557} & \underline{0.522} & 0.487 & \underline{0.462} & \textbf{0.532} & \textbf{0.537}\\
\midrule
\multicolumn{11}{@{}l}{\textit{DeepSeek-V4-Flash-0731}}\\
Base LLM & 0.368 & 0.230 & 0.236 & 0.257 & 0.274 & 0.232 & 0.246 & 0.265 & 0.224 & 0.266\\
Full context & 0.533 & 0.512 & 0.514 & 0.512 & 0.553 & 0.545 & 0.529 & 0.516 & 0.524 & 0.527\\
\cmidrule(l){1-11}
RAG & 0.468 & 0.371 & 0.343 & 0.396 & 0.410 & 0.372 & 0.355 & \underline{0.402} & 0.361 & 0.391\\
Rewrite & \underline{0.513} & \underline{0.419} & \underline{0.440} & 0.401 & \underline{0.419} & 0.415 & \textbf{0.453} & 0.394 & \underline{0.441} & \underline{0.438}\\
Mem0 & 0.450 & 0.412 & 0.387 & 0.379 & 0.402 & \underline{0.439} & \underline{0.433} & 0.373 & 0.415 & 0.412\\
A-MEM & 0.501 & 0.367 & 0.363 & \underline{0.416} & 0.397 & 0.388 & 0.413 & 0.386 & 0.388 & 0.408\\
MemListener & 0.415 & 0.357 & 0.383 & 0.338 & 0.402 & 0.350 & 0.405 & 0.335 & 0.374 & 0.376\\
\rowcolor{qabmblue}HabitusMem & \textbf{0.646} & \textbf{0.430} & \textbf{0.442} & \textbf{0.471} & \textbf{0.493} & \textbf{0.478} & \underline{0.435} & \textbf{0.402} & \textbf{0.466} & \textbf{0.484}\\
\bottomrule
\end{tabular}
\end{table*}

\paragraph{Personalized answering.}
Table~\ref{tab:pm-baselines} shows that HabitusMem is the most accurate memory method under both agents, at 0.537 with Qwen and 0.484 with DeepSeek, 4.8 and 4.6 points above the strongest baseline. That baseline changes with the agent and the scenario, whereas HabitusMem leads in five of nine scenarios with Qwen and eight with DeepSeek, so belief memory transfers more uniformly than content-based memory. Conditioning on the request does not explain this advantage by itself, since RAG also retrieves for the current request yet reaches only 0.460 and 0.391. Retrieval returns what the user said that resembles the request, whereas a belief record states what the history implies about the user for that request. In the Knowledge scenario, the memory-free base LLM with Qwen outperforms full context and every content-based memory, indicating that injected history interferes with questions the agent could otherwise answer, whereas HabitusMem conveys only the relevant beliefs and exceeds full context by 13 and 11 points. Full context remains more accurate, as expected of a condition that loses no evidence, but the generation score shows what evidence alone does not provide. Moving from the memory-free base LLM to full context raises accuracy with Qwen and DeepSeek from 0.446 and 0.263 to 0.569 and 0.522, yet generation stays nearly flat, from 0.256 and 0.291 to 0.227 and 0.302, while HabitusMem reaches 0.326 and 0.382. Answer options likely cue the agent toward the relevant preference, which it can then recognize in the raw history, whereas an open answer requires the agent to infer and apply that preference unaided. Belief memory performs this inference before the agent acts, which is the motivation for representing the user rather than the history.

\begin{table*}[!t]
\centering\footnotesize
\caption{Personalized tool-use results on VitaBench 2.0. Bold and underlined values mark the best and second-best compact memory systems in each column.}
\label{tab:vb}
\begin{tabular}{@{}l*{6}{w{c}{36pt}}@{}}
\toprule
\multirow{2}{*}{Memory} & \multicolumn{3}{c}{Qwen3.5-397B-A17B Agent} & \multicolumn{3}{c}{DeepSeek-V4-Flash-0731 Agent}\\
\cmidrule(lr){2-4}\cmidrule(l){5-7}
& $\avgfour$ & $\passany$ & $\passall$ & $\avgfour$ & $\passany$ & $\passall$\\
\midrule
Base LLM & 0.0895 & 0.2010 & 0.0182 & 0.0833 & 0.1946 & 0.0130\\
Full context & 0.3382 & 0.5888 & 0.1297 & 0.3119 & 0.5409 & 0.1102\\
Ground-truth profile & 0.3330 & 0.5888 & 0.1193 & 0.3839 & 0.6316 & 0.1569\\
\midrule
RAG & 0.2857 & 0.5391 & 0.0864 & 0.2251 & 0.4705 & 0.0562\\
Rewrite & \underline{0.3139} & \textbf{0.6109} & 0.0700 & 0.3123 & \underline{0.5863} & 0.0869\\
Mem0 & 0.2844 & \underline{0.5694} & 0.0636 & 0.2643 & 0.5707 & 0.0480\\
A-MEM & 0.2964 & 0.5486 & \underline{0.0999} & \underline{0.3165} & \textbf{0.5875} & 0.0843\\
MemListener & 0.2202 & 0.4189 & 0.0687 & 0.2422 & 0.4293 & \underline{0.0934}\\
\rowcolor{qabmblue}HabitusMem & \textbf{0.3291} & 0.5616 & \textbf{0.1310} & \textbf{0.3573} & 0.5798 & \textbf{0.1284}\\
\bottomrule
\end{tabular}
\end{table*}

\paragraph{Personalized tool use.}
Table~\ref{tab:vb} shows that HabitusMem obtains the highest average reward among memory methods under both agents and the highest all-success of any non-privileged condition, 0.1310 and 0.1284 against 0.1297 and 0.1102 for full context. The results reveal a gap between coverage and consistency. Rewrite succeeds at least once on 61\% of subtasks with Qwen but in all four trials on only 7\%, whereas among subtasks solved at least once, HabitusMem succeeds in every trial on 23\% with Qwen and 22\% with DeepSeek, compared with 22\% and 20\% for full context. A history summary preserves much of what happened but leaves the agent to reconstruct the user's choice on every run, so success varies from trial to trial, whereas a belief record that states the choice and its constraints fixes the decision across runs. The privileged ground-truth profile further shows that correct preferences alone do not secure execution. With Qwen it reaches 0.1193 all-success, below 0.1310 for HabitusMem, likely because a static profile does not carry the exact choices and constraints of a specific transaction.

\subsection{Component analysis}
\label{sec:components}
\begin{figure}[t]
\centering
\includegraphics[width=\textwidth]{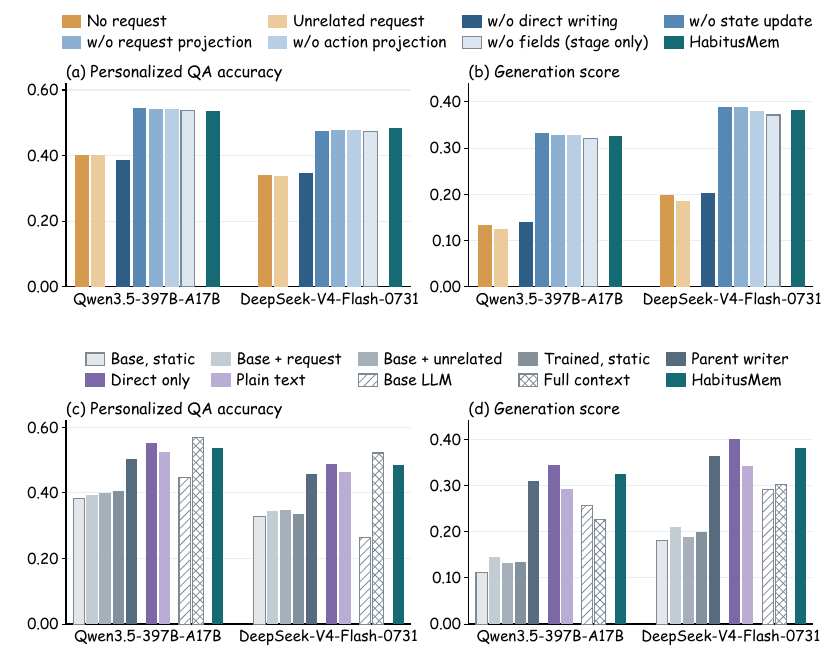}
\caption{\textbf{Component analysis on PersonaMem-v2.} (a--b) Inference interventions with a fixed writer and leave-one-out retraining. (c--d) Training lineage, matched controls, and reference conditions. Bars are grouped by agent, the rightmost bar in each group is HabitusMem, and hatched bars are reference conditions. Tables~\ref{tab:component-ablation} and~\ref{tab:analysis} list exact values.}
\label{fig:components}
\end{figure}

\begin{table}[t]
\centering\footnotesize
\setlength{\tabcolsep}{4pt}
\caption{Training lineage and matched-control results on VitaBench 2.0.}
\label{tab:vb-controls}
\begin{tabular}{@{}l*{6}{w{c}{38pt}}@{}}
\toprule
& \multicolumn{3}{c}{Qwen3.5-397B-A17B Agent} & \multicolumn{3}{c}{DeepSeek-V4-Flash-0731 Agent}\\
\cmidrule(lr){2-4}\cmidrule(l){5-7}
Writer & $\avgfour$ & $\passany$ & $\passall$ & $\avgfour$ & $\passany$ & $\passall$\\
\midrule
Base, static & 0.2510 & 0.4620 & 0.0840 & 0.2700 & 0.5040 & 0.0840\\
Base + request & 0.3410 & 0.5380 & 0.1620 & 0.2897 & 0.5320 & 0.0866\\
Base + unrelated & 0.1940 & 0.3700 & 0.0610 & 0.2100 & 0.4250 & 0.0600\\
Trained, static & 0.2640 & 0.4530 & 0.1040 & 0.2480 & 0.4660 & 0.0890\\
Parent writer & 0.3012 & 0.5214 & 0.1245 & 0.3243 & 0.5499 & 0.1310\\
\midrule
Direct only & 0.2944 & 0.4929 & 0.1284 & 0.3136 & 0.5227 & 0.1349\\
Plain text & 0.3252 & 0.4929 & 0.1725 & 0.3324 & 0.5344 & 0.1362\\
\rowcolor{qabmblue}HabitusMem & 0.3291 & 0.5616 & 0.1310 & 0.3573 & 0.5798 & 0.1284\\
\bottomrule
\end{tabular}
\end{table}

Figure~\ref{fig:components} and Table~\ref{tab:vb-controls} report the contribution of each design component on PersonaMem-v2 and VitaBench 2.0, and Appendix~\ref{app:component-protocols} gives the complete protocols. Paired PersonaMem-v2 scores list Qwen first and DeepSeek second.

\paragraph{Request conditioning.}
Two inference interventions test whether the writer depends on the request. Both keep the trained adapter, instruction, history selection, and generation settings fixed and change only the request passed to the writer. \emph{Without request conditioning} removes the request suffix. \emph{With an unrelated request} keeps the suffix format and replaces its content with one fixed question for every query. Both interventions lower accuracy from 0.537 and 0.484 to about 0.40 and 0.34, and generation from 0.326 and 0.382 to 0.12--0.20. The two interventions have nearly the same effect, which shows that the gain comes from the content of the request and not from the suffix format.

The training lineage tests whether the writer must learn to use the request. \emph{Base, static} is the unadapted Qwen3.5-9B writer with the belief instruction and no request. \emph{Base + request} adds the current request to this writer, and \emph{Base + unrelated} replaces it with the fixed unrelated question. \emph{Trained, static} is the cold-start writer evaluated without a request. The \emph{Parent writer} is the direct-writing writer that initializes HabitusMem. Base, static reaches 0.383 and 0.329. The current request raises accuracy to 0.393 and 0.345, and the unrelated question raises it to 0.401 and 0.347, so the content of the request does not improve the accuracy of the unadapted writer. Trained, static reaches 0.406 and 0.335. The request alone and training alone therefore each add at most 2.3 points. The Parent writer, which combines training with the request, reaches 0.505 and 0.457, and HabitusMem reaches 0.537 and 0.484. Section~\ref{sec:formulation} states that conditioning on the request can only lower the best attainable risk under a fixed budget, and these results show that a writer must learn to attain this benefit.

The request also steers the unadapted writer on VitaBench 2.0. In the archived Qwen runs, Base + request reaches 0.3410 average reward, Base, static reaches 0.2510, and Base + unrelated reaches 0.1940. With DeepSeek, the three conditions reach 0.2897, 0.2700, and 0.2100. An unrelated request lowers average reward below the no-request condition under both agents, likely because it directs the memory budget toward irrelevant content.

\paragraph{Training tasks.}
Each leave-one-out variant removes all examples of one task from the 40,766-row four-task training file and is retrained from the parent with the same recipe, so the removed task is the only difference. \emph{Leave out fields (stage only)} keeps all rows and removes the record schema from the instruction and targets of the final stage. Removing direct writing lowers accuracy to 0.387 and 0.348, the level of the untrained writers, although the other three tasks remain. Direct writing is the only task that trains the deployed mapping from a raw history and a request to records, because state update receives no request and the projection tasks start from existing records.

\emph{Direct only} retrains the parent on the 18,529 direct-writing examples alone for 5,096 steps, which matches the optimization budget of the full mixture. On tool use, the full mixture improves paired average reward over Direct only by 3.3 points with Qwen and 4.5 points with DeepSeek, and it raises Qwen all-success over the Parent writer from 0.1245 to 0.1310. This result is consistent with the content of the training targets. Direct-writing targets place the annotated preference at the top of the memory, and action-projection targets require the exact choices and fulfillment constraints that a transaction needs. The leave-one-out runs were evaluated on answering only, so this result applies to the auxiliary tasks as a group and not to any single task.

\paragraph{Record fields.}
\emph{Plain text} repeats the cold-start and four-task stages from the base model with the same rows and settings. It converts the instruction and targets to plain bullets by removing the record header and the evidence, scope, and status fields. On PersonaMem-v2, the full format adds 1.3 and 1.9 accuracy points and 3.3 and 3.9 generation points over Plain text. Appendix~\ref{app:probes} reports controlled probes of these fields.

\section{Conclusion}
\label{sec:conclusion}
We treated personalized memory as the assistant's belief about an evolving user state, conveyed as the part relevant to the current decision with the evidence, scope, and status of each claim. With one writer shared across benchmarks and two agent families, HabitusMem is the most accurate evaluated compact memory on PersonaMem-v2 and achieves the highest all-success among non-privileged VitaBench conditions. The component analysis separates the sources of these results: learned request conditioning is central to answering, the auxiliary tasks improve tool use, and structured fields provide an answering gain.

\paragraph{Limitations.}
HabitusMem has three main limitations. First, its fixed input budget may exclude relevant evidence from long interaction histories. Second, belief extraction and application remain imperfect, particularly for retracted preferences and precise execution constraints. Third, we have not yet evaluated HabitusMem in broader applications, including email composition, personalized recommendations, and scheduling.

\subsection*{Reproducibility Statement}
Section~\ref{sec:method} describes the writer interface and supervision. Section~\ref{sec:setup} specifies the evaluation protocol, and Section~\ref{sec:components} reports the component comparisons. Appendix~\ref{app:implementation} gives training, evaluation, and statistical details; Appendices~\ref{app:sensitivity} and~\ref{app:memory-analysis} contain sensitivity, cost, and probe results. Appendix~\ref{app:cases} gives case studies.
\clearpage
\bibliographystyle{plainnat}
\bibliography{main}
\clearpage
\appendix
\section{Decision-theoretic interpretation of belief memory}
\label{app:bayesian}
Section~\ref{sec:formulation} models the user as an evolving latent state and defines memory by expected loss under the belief, following Bayesian decision theory~\citep{berger1985statistical}: the decision variable is a memory message, and its value depends on the actions it induces in a fixed agent. The formulation is a design rationale for what a record should state, and HabitusMem does not carry out the inference numerically. This appendix separates the evidential and decisional roles of the request, relates the formulation to habitus, and shows when conditioning the memory on the request helps.

\subsection{The request as evidence and as decision}
Let $\bar b_t(z_t)=\int p(z_t\mid z_{t-1})\,b_{t-1}(z_{t-1})\,dz_{t-1}$ denote the belief predicted from the interactions before the request. Setting $o_t=q_t$ in Equation~\ref{eq:posterior} gives
\begin{equation}
b_t(z_t)\propto p(q_t\mid z_t)\,\bar b_t(z_t).
\label{eq:query-evidence}
\end{equation}
If the request carries no evidence about the user, its likelihood is constant in $z_t$ and the belief equals the prediction from earlier interactions. The request still enters the loss $\ell_\tau(a,z_t;q_t)$ and can therefore change the ideal memory: a gardening preference can become central to an activity recommendation without becoming more credible because the user asks for an activity. A request that reveals a new circumstance, such as a changed delivery address, changes both the belief and the loss. In the implementation, input selection additionally determines which part of the history the writer sees.

\subsection{Habitus and the latent state}
Bourdieu describes habitus as durable dispositions, produced by past experience, that generate practices in concrete situations \citep{bourdieu1977outline}, and as an open system of dispositions that new experiences either reinforce or modify \citep[p.~133]{bourdieu1992invitation}. In our formulation, the interaction history is the record of practices from which the assistant infers the user's state, the transition $p(z_t\mid z_{t-1})$ allows that state to change, and the scope field records the situations in which a disposition guides behavior. The computational state $z_t$ is broader than habitus, since it also includes facts and explicit constraints that are not dispositions. The belief represents the assistant's uncertainty about the user rather than a model of the user's own reasoning; habitus motivates the formulation, and the experiments evaluate the memory design that follows from it.

\subsection{The value of conditioning memory on the request}
When memories must fit a message budget, conditioning on the request can only lower the best attainable risk. Fix a common history selection $\bar H=T(H)$, a task type $\tau$, the agent and its guidance, and a budget $B$, and let $\mathcal M_B\subseteq\mathcal M$ contain the memories of at most $B$ tokens. Write $\mathcal R_\tau(m;\bar H,Q)$ for the risk in Equation~\ref{eq:objective} given the selected history $\bar H$ and the request $Q$. Let $\mathcal C_B$ contain all encoders $c:\bar H\mapsto\mathcal M_B$, and let $\mathcal W_B$ contain all encoders $w:(\bar H,Q)\mapsto\mathcal M_B$. Assuming the expected risks exist, the best attainable decision risk satisfies
\begin{equation}
\inf_{w\in\mathcal W_B}\mathbb E_{\bar H,Q}
\left[\mathcal R_\tau(w(\bar H,Q);\bar H,Q)\right]
\le
\inf_{c\in\mathcal C_B}\mathbb E_{\bar H,Q}
\left[\mathcal R_\tau(c(\bar H);\bar H,Q)\right].
\label{eq:conditioning-observation}
\end{equation}
\paragraph{Proof.}
For every $c\in\mathcal C_B$, the encoder $w_c(h,q)=c(h)$ belongs to $\mathcal W_B$ and produces the same memory and risk. Taking the infimum over the larger class gives the inequality. The comparison holds the information, agent, and output budget fixed and concerns the best attainable risk; how closely a trained writer approaches it is an empirical question, which Section~\ref{sec:components} examines.

A one-bit example shows that the inequality can be strict. Suppose the history contains independent uniform bits $(U,V)$ and a uniform request asks for one of them. A request-conditioned encoder sends the requested bit and achieves zero error with a one-bit message. A static one-bit encoder, even with an optimal decoder, has minimum average error $1/4$: the four histories require four answer pairs, but one bit permits only two reconstruction pairs. At least two of the eight equally likely history--request cases are therefore answered incorrectly, and sending $U$ attains the bound. The example isolates the benefit of allocating a limited message to a known request.

\subsection{Textual records and the learned approximation}
A record $(\text{claim},\text{evidence},\text{scope},\text{status})$ communicates a user-related assertion together with its source, applicability, and current standing. Its evidence field refers to an observation already present in the history and is not an independent observation in Equation~\ref{eq:posterior}. Likewise, the token probabilities in Equation~\ref{eq:sft} model textual continuations, not calibrated probabilities that the claims are true. The records therefore express the belief qualitatively, through the stated evidence and status of each claim, and not as posterior probabilities.

The status field distinguishes changes that a truth probability alone cannot express. A correction changes what the assistant believes happened, a preference update changes what is currently desired, and a request to stop using information restricts its use without necessarily making it false. Explicit records make these distinctions available to the agent, and correct use depends on both the generated memory and the agent. Action-oriented supervision likewise consists of demonstrated textual memories whose effects are evaluated on downstream agent tasks; it does not train on observed user actions.

\section{Implementation and evaluation details}
\label{app:implementation}
\subsection{Inference and history selection}
HabitusMem uses the same adapter on both benchmarks, with the inference settings of Section~\ref{sec:construction}. Both evaluations use Hugging Face generation with greedy decoding and a serialized generation lock. The writer regenerates the memory whenever the history--request pair changes, so the memory can change during a VitaBench interaction without recursive updates to a persistent state; state-update and projection examples serve as training supervision only. The VitaBench selector reserves space for the instruction, the request suffix, and 32 additional tokens, and then combines an initial history segment, an elision marker, and the most recent interactions. The PersonaMem selector keeps the first 32,000 characters before the final token cutoff.

\subsection{Training and evaluation settings}
\label{app:training}
The LoRA adapter has rank 64, alpha 128, and dropout 0.05. The cold-start stage trains for two epochs at learning rate $5\times10^{-5}$ on 1,598 belief memories, which the DeepSeek-V4-Flash teacher writes at two history positions for each of the 799 training personas. The parent stage trains for one epoch on 18,529 direct-writing examples. The four-task stage trains for one epoch with a cosine schedule on 40,766 rows, comprising 18,529 direct-writing, 6,270 state-update, 6,379 request-projection, and 9,588 action-projection rows, the last of which repeat 3,196 examples three times. The parent and four-task stages both use learning rate $2\times10^{-5}$, a 12,288-token sequence cutoff, and an effective batch of 8. The base checkpoint is a vision-language model, and all inputs in our experiments are text. LoRA targets \code{q\_proj}, \code{k\_proj}, \code{v\_proj}, \code{o\_proj}, the linear-attention \code{in\_proj\_*} modules, and the feed-forward \code{gate\_proj}, \code{up\_proj}, and \code{down\_proj}. Because the targeting rule matches all linear layers, 32.12M of the 205.23M adapter parameters (15.65\%) go to visual projection and feed-forward modules, which text inputs do not use. Embeddings and the output head are not trained.

Each direct-writing example pairs the instruction, the selected history, and the current request with an annotated memory. The record inserted at the top of the static memory, under the \code{\#\# Beliefs} header, has one of three forms:
\begin{tcolorbox}[colback=qabmblue!45,colframe=qabmteal,boxrule=0.5pt,arc=2mm,left=2mm,right=2mm,top=1mm,bottom=1.5mm,before skip=6pt,after skip=8pt]
\small\raggedright
\textcolor{qabmteal}{\textbf{Active}}\par
\code{- \{preference\} | evidence: user statement | scope: always | status: active}\par\medskip
\textcolor{qabmgold}{\textbf{Update}}\par
\code{- [update] \{preference\} | evidence: latest statement | scope: always | status: active (supersedes the earlier preference on this topic)}\par\medskip
\textcolor{qabmred}{\textbf{Retraction}}\par
\code{- The user has asked to forget a preference regarding \{topic\}; do not use or mention it | evidence: user instruction | scope: always | status: retracted}
\end{tcolorbox}

Auxiliary targets use 5,000-token history chunks, an 11,400-token input budget, teacher generation caps of 4,800 tokens for states and 1,800 for views, and final state and view budgets of 2,400 and 800 tokens. The data builder merges records using line-level deduplication and ordering heuristics, and the parent writes up to eight request-conditioned and four action-conditioned views per persona. Target generation is retried when no syntactically complete record is produced. All examples come from the official training split, and direct writing uses its preference annotations as target supervision. Action-projection examples take teacher-constructed states as inputs, whereas VitaBench inference takes raw history, so the training prompts for this task differ from the deployment prompt.

The ground-truth profile is a privileged reference and full context exposes the history directly. All VitaBench systems share the same execution guidance and four-trial protocol. PersonaMem-v2 uses the official training and test splits, with all 5,000 official test queries evaluated. Component results are reported in Section~\ref{sec:components}.

Tables~\ref{tab:component-ablation} and~\ref{tab:analysis} list the exact PersonaMem-v2 scores plotted in Figure~\ref{fig:components}.

\begin{table}[t]
\centering\footnotesize
\setlength{\tabcolsep}{4pt}
\caption{Request and training ablations on PersonaMem-v2.}
\label{tab:component-ablation}
\begin{tabular}{@{}l*{4}{w{c}{46pt}}@{}}
\toprule
& \multicolumn{2}{c}{Qwen3.5-397B-A17B} & \multicolumn{2}{c}{DeepSeek-V4-Flash-0731}\\
\cmidrule(lr){2-3}\cmidrule(l){4-5}
Setting & Accuracy & Generation & Accuracy & Generation\\
\midrule
\rowcolor{qabmblue}HabitusMem (default) & 0.5372 & 0.3257 & 0.4842 & 0.3824\\
\midrule
\multicolumn{5}{@{}l}{\textit{Inference intervention (writer fixed)}}\\
Without request conditioning & 0.4004 & 0.1345 & 0.3394 & 0.1995\\
With an unrelated request & 0.4019 & 0.1247 & 0.3376 & 0.1859\\
\midrule
\multicolumn{5}{@{}l}{\textit{Training component removed (retrained)}}\\
Leave out direct writing & 0.3871 & 0.1411 & 0.3476 & 0.2026\\
Leave out state update & 0.5444 & 0.3333 & 0.4734 & 0.3900\\
Leave out request projection & 0.5429 & 0.3283 & 0.4775 & 0.3899\\
Leave out action projection & 0.5425 & 0.3281 & 0.4786 & 0.3799\\
Leave out fields (stage only) & 0.5383 & 0.3202 & 0.4724 & 0.3717\\
\bottomrule
\end{tabular}
\end{table}

\begin{table}[t]
\centering\footnotesize
\setlength{\tabcolsep}{4pt}
\caption{Training lineage and matched answering controls on PersonaMem-v2.}
\label{tab:analysis}
\begin{tabular}{@{}l*{4}{w{c}{46pt}}@{}}
\toprule
& \multicolumn{2}{c}{Qwen3.5-397B-A17B} & \multicolumn{2}{c}{DeepSeek-V4-Flash-0731}\\
\cmidrule(lr){2-3}\cmidrule(l){4-5}
Writer & Accuracy & Generation & Accuracy & Generation\\
\midrule
\multicolumn{5}{@{}l}{\textit{Training lineage}}\\
Base, static & 0.383 & 0.112 & 0.329 & 0.181\\
Base + request & 0.393 & 0.146 & 0.345 & 0.211\\
Base + unrelated & 0.401 & 0.133 & 0.347 & 0.189\\
Trained, static & 0.406 & 0.135 & 0.335 & 0.200\\
Parent writer & 0.505 & 0.310 & 0.457 & 0.364\\
\midrule
\multicolumn{5}{@{}l}{\textit{Matched controls}}\\
Direct only & 0.553 & 0.345 & 0.487 & 0.401\\
Plain text & 0.524 & 0.293 & 0.465 & 0.343\\
\rowcolor{qabmblue}HabitusMem & 0.537 & 0.326 & 0.484 & 0.382\\
\midrule
\multicolumn{5}{@{}l}{\textit{Reference conditions}}\\
Base LLM & 0.446 & 0.256 & 0.263 & 0.291\\
Full context & 0.569 & 0.227 & 0.522 & 0.302\\
\bottomrule
\end{tabular}
\end{table}

\subsection{Component-analysis protocols}
\label{app:component-protocols}
All component conditions are evaluated on the full 5,000-query PersonaMem-v2 benchmark with the same agents and judge. The training-lineage conditions and matched controls are also evaluated on VitaBench 2.0 with the same agents. The inference interventions, lineage conditions, and matched controls otherwise use the final inference interface; only the factor named in each condition changes.

\paragraph{Inference interventions.}
Both interventions keep the trained adapter, instruction, selected history, and generation settings fixed and change only the request passed to the writer. \emph{Without request conditioning} removes the request suffix, so the writer sees the instruction and history alone. \emph{With an unrelated request} retains the suffix format used in training but replaces its content with the fixed question ``Recommend a good book on European history for a beginner'' for every query. Memories are regenerated under each condition, and the same queries are re-answered by the same agent and judge, so the comparison is paired at the query level.

\paragraph{Leave-one-out training components.}
The final training file contains 18,529 direct-writing, 6,270 state-update, 6,379 request-projection, and 3,196 action-projection examples; action-projection examples are repeated three times, giving 40,766 rows and weights $(1,1,1,3)$. Each leave-one-out variant is a filtered copy of this file that removes all rows of one task and keeps the other three, with the same repetition counts. Every variant is retrained from the parent writer with the same LoRA rank, alpha, dropout, learning rate, schedule, cutoff, and one-epoch budget as the default, so the dropped task is the only difference. The variants contain 22,237 rows without direct writing, 34,496 without state update, 34,387 without request projection, and 31,178 without action projection. \emph{Leave out fields (stage only)} keeps all 40,766 rows but replaces the instruction with a schema-free instruction and strips the \code{evidence}, \code{scope}, and \code{status} suffixes from the targets; it starts from the same schema-trained parent, so only the final stage is trained on plain records.

\paragraph{Training lineage.}
\emph{Base, static} is the unadapted Qwen3.5-9B writer with the belief instruction and no request suffix. \emph{Base + request} adds the current request to the same unadapted writer, and \emph{Base + unrelated} keeps the suffix but replaces its content with the fixed unrelated question. \emph{Trained, static} is the cold-start writer, trained for two epochs on 1,598 teacher-written belief memories, evaluated without request conditioning. The \emph{Parent writer} is the direct-writing writer trained for one epoch on the 18,529 request-conditioned examples of the official training split; it initializes the final writer and is the teacher that generates the state-update and projection targets. All five conditions use the final inference interface and the same test set.

\paragraph{Matched controls.}
The direct-only control is retrained from the same parent with the same recipe on the 18,529 direct-writing examples alone. It runs for 2.2 epochs, matching the 5,096 steps of the 40,766-row full mixture, so the comparison holds the optimization budget fixed while removing the auxiliary tasks. The plain-text control starts from the base model and repeats the cold-start and four-task stages with the same examples, queries, annotated preferences, selector, output cap, and optimization settings: it trains a plain-text cold-start writer for two epochs on the 1,598 cold-start memories at learning rate $5\times10^{-5}$ and then a four-task writer for one epoch on the 40,766-row mixture at $2\times10^{-5}$. The instruction and targets are converted mechanically: the \code{\#\# Beliefs} header and the \code{evidence}, \code{scope}, and \code{status} fields are removed without paraphrasing, and the instruction asks for plain bullets. The targets are text transforms of the final training file, and no teacher is rerun.

Paired PersonaMem-v2 accuracy comparisons use query-level correctness and McNemar tests. VitaBench average reward uses paired bootstrap over task clusters, while any-success and all-success use McNemar tests over common completed subtasks. Because these tests use common subtasks, the full-run aggregates can differ from the means on paired subsets. Each writer is trained once, so these tests do not capture variation across training seeds. Qwen training-lineage results use archived full runs of the same Qwen3.5-397B-A17B agent.

For the direct-only comparison, the full mixture improves paired average reward by 3.3 points with Qwen and 4.5 with DeepSeek. It also improves Qwen all-success from 0.1284 to 0.1310 and DeepSeek any-success from 0.5227 to 0.5798. For the parent comparison, the Qwen all-success gain is concentrated in the 668 personalization subtasks. Its paired average-reward gain is 2.6 points. The DeepSeek parent comparison has a paired reward gain of 2.8 points with an interval of $[+0.6,+4.9]$ points. For the plain-text comparison, DeepSeek's paired reward difference is 1.8 points, and any-success and all-success differ by 3.6 and 0.9 points, none of which is statistically significant.

\section{Parameter sensitivity and computational cost}
\label{app:sensitivity}
\subsection{Training-mixture sensitivity}
\label{app:mixture-sensitivity}
\begin{figure}[t]
\centering
\includegraphics[width=\textwidth]{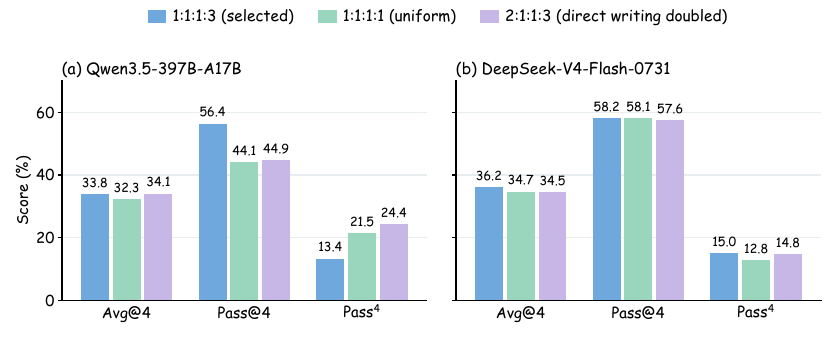}
\caption{Sensitivity to training-mixture weights on the full VitaBench 2.0 benchmark. Weights follow direct writing, state update, request projection, and action projection. Scores are percentages.}
\label{fig:mixture-sensitivity}
\end{figure}

We did not run an extensive hyperparameter search because the purpose of this comparison is to test sensitivity to the weights rather than to optimize them. We therefore compare three common training mixtures. We train three writers from the same parent with the identical recipe (one epoch, learning rate $2\times10^{-5}$, LoRA rank 64, $\alpha=128$), changing only the repetition counts in the training file. The selected $(1,1,1,3)$ mixture repeats each action-projection example three times, the uniform $(1,1,1,1)$ mixture repeats every example once, and the direct-doubled $(2,1,1,3)$ mixture repeats direct-writing examples twice while keeping action projection at three. Figure~\ref{fig:mixture-sensitivity} compares the three mixtures.

On the full DeepSeek benchmark, selected reaches 0.3619 average reward and 0.1505 all-success, against 0.3469 and 0.1284 for uniform and 0.3450 and 0.1479 for direct-doubled. Their any-success values are 0.5823, 0.5811, and 0.5759. With Qwen, average reward is indistinguishable (0.3379, 0.3233, 0.3405), but selected has higher any-success (0.5642 against 0.4410 and 0.4488) and lower all-success (0.1336 against 0.2153 and 0.2438). Uniform and direct-doubled do not differ significantly on any metric. The selected and uniform writers both reach 0.4842 PersonaMem accuracy. Thus, weight selection does not yield a uniform advantage across agents and metrics.

\subsection{Inference sensitivity}
\label{app:inference-sensitivity}
\begin{figure}[t]
\centering
\includegraphics[width=\textwidth]{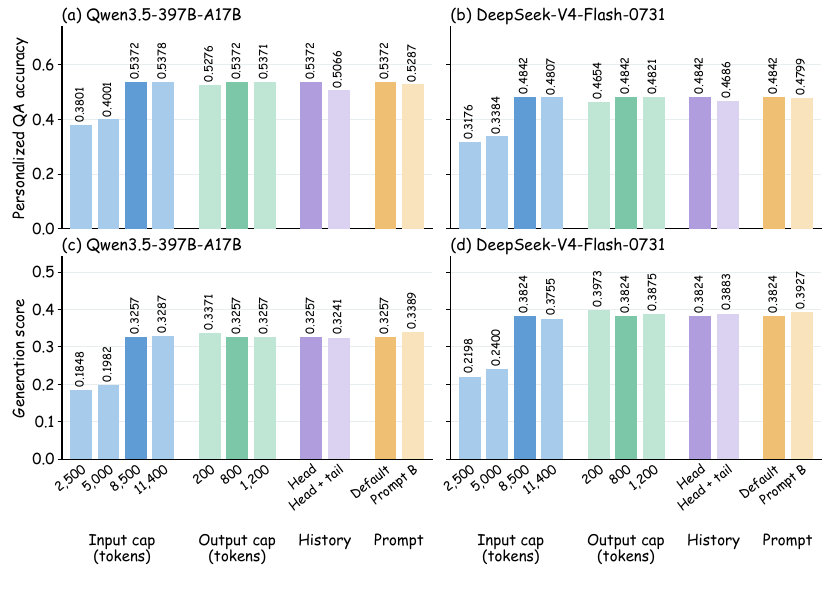}
\caption{\textbf{Inference sensitivity on PersonaMem-v2.} (a--b) Personalized QA accuracy and (c--d) generation score. Each cluster varies one setting while the others stay at their defaults, and darker bars mark the default configuration.}
\label{fig:inference-sensitivity}
\end{figure}

The trained writer is fixed, and all PersonaMem-v2 queries are answered under each setting with both agents. We vary one setting at a time: input caps of 2,500, 5,000, 8,500 (default), and 11,400 tokens; output caps of 200, 800 (default), and 1,200 tokens; a head-and-tail selector that reserves 15\% of the input token budget for the beginning of the history and the rest for the most recent interactions; and Prompt B, an alternative instruction that keeps the record schema but prioritizes concrete preferences, facts, and constraints and limits biographical background to two lines. Figure~\ref{fig:inference-sensitivity} reports the results. Reducing the input cap to 2,500 or 5,000 tokens costs 13.7--16.7 accuracy points and 12.8--16.3 generation points across the two agents. Raising it to 11,400 changes accuracy by at most 0.4 points and generation by at most 0.7. Head-and-tail selection incurs the largest accuracy loss among the other choices (3.06 points with Qwen), with little generation change. A 200-token output cap slightly reduces accuracy but improves generation under both agents; expanding to 1,200 tokens yields little benefit. Prompt B similarly trades a small accuracy decrease for a generation increase. These effects are much smaller than those of removing relevant history or request conditioning.

\subsection{Computational cost}
\label{app:cost}
The four-task training stage takes 76.3 GPU-hours. In the PersonaMem-v2 32k setting, the writer reads a median of 7,125 tokens and generates 443 tokens per request. Evaluating 5,000 distinct requests takes roughly 24 writer-hours on one GPU. The memory presented to the agent has a median length of 515 tokens, compared with 31,855 tokens for full history, a 61.9-fold reduction in history input. An unchanged history--request pair can reuse its cached memory; a new history or request requires another writer pass.

\section{Memory quality and agent behavior}
\label{app:memory-analysis}
Belief memory exposes not only a claim, but also its evidence, applicability, and current status. We design diagnostics around these distinctions to examine two separate steps: whether the writer preserves the relevant fact and whether a frozen agent uses it in a decision. A constructed paired probe suite tests these steps under controlled history changes.

\subsection{Constructed probes of evidence, scope, and status}
\label{app:probes}
We construct 90 probes per writer across ten topics and nine cells: a self-purchase and gift pair for attribution, unconditional and conditional preferences crossed with partner and solo questions for scope, and active, retracted, and superseded preferences for status. Each topic therefore contributes two attribution, four scope, and three status probes. Within each history pair or triple, the question and four choices stay fixed while one history element changes. The full writer is evaluated with both agents; the matched plain-text writer is evaluated with Qwen. The suite is constructed rather than sampled from either benchmark because this design requires controlled alternatives for the same decision. The writer, agents, and inference configurations otherwise match the benchmark evaluations. The writer sees the history and question; the answering agent sees the generated memory and question without the raw history. A separate call to the agent model checks whether the memory states the manipulated fact, and we cross-check those judgments by inspecting the records. This design makes the fields operational: it tests whether a change in evidence, scope, or status survives the memory bottleneck and changes the downstream answer.

\paragraph{A concrete attribution pair.}
The two histories of the jasmine-tea pair differ only in their third turn. The question and the four choices are the same in both conditions.
\begin{tcolorbox}[colback=qabmblue!45,colframe=qabmteal,boxrule=0.5pt,arc=2mm,left=2mm,right=2mm,top=1mm,bottom=1.5mm,before skip=6pt,after skip=8pt]
\small
\textcolor{qabmteal}{\textbf{Shared history}}\par
The opening user statement ``I've really gotten into jasmine tea lately,'' the assistant's acknowledgment, and a final unrelated turn about documentaries.\par\medskip
\textcolor{qabmgold}{\textbf{Third turn}}\par
\textbf{Self:} ``I picked up jasmine tea again yesterday and loved it.''\par
\textbf{Gift:} ``I picked up jasmine tea yesterday---it's a birthday gift for my sister, she loves it.''\par\medskip
\textcolor{qabmteal}{\textbf{Question}}\par
``I want to pick something up for myself this weekend---what would you suggest I go for?''\par\medskip
\textcolor{qabmteal}{\textbf{Choices}}\par
\textbf{A.} Since you're into jasmine tea, go for that again.\par
\textbf{B.} Try something completely new that you haven't mentioned before.\par
\textbf{C.} Pick whatever is on sale, brands don't matter.\par
\textbf{D.} Ask a friend what they would choose for you.\par\medskip
\textcolor{qabmred}{\textbf{Answer key}}\par
A for the self condition and B for the gift condition.
\end{tcolorbox}
The attribution axis contains ten such pairs, one per topic, so each of these two cells contains ten probes.

\begin{table}[!htbp]
\centering\footnotesize
\setlength{\tabcolsep}{4pt}
\caption{\textbf{All controlled probe results.} Entries are correct agent answers out of ten. The gift cell also includes an earlier self-statement of interest; it does not isolate writer attribution. Manual inspection finds the gift relationship in all ten memories from both writers.}
\label{tab:probes}
\begin{tabular}{@{}llccc@{}}
\toprule
Factor & Condition & Full, Qwen & Full, DeepSeek & Plain, Qwen\\
\midrule
Attribution & Self & 10/10 & 10/10 & 10/10\\
 & Gift / prior interest & 0/10 & 1/10 & 0/10\\
\midrule
Scope & Unconditional, partner & 10/10 & 10/10 & 10/10\\
 & Unconditional, solo & 10/10 & 10/10 & 10/10\\
 & Conditional, partner & 10/10 & 10/10 & 10/10\\
 & Conditional, solo & 10/10 & 2/10 & 7/10\\
\midrule
Status & Active & 10/10 & 10/10 & 10/10\\
 & Retracted & 1/10 & 2/10 & 0/10\\
 & Superseded & 9/10 & 9/10 & 8/10\\
\bottomrule
\end{tabular}
\end{table}

\paragraph{What the fields preserve and what the agent applies.}
Table~\ref{tab:probes} separates record quality from downstream answers and shows that explicit fields preserve distinctions that plain records lose. The full writer records the conditional scope and the retraction in all ten corresponding memories and the supersession in nine to ten, whereas the matched plain-text writer marks no retractions. With Qwen, full records match or exceed plain records in every cell. They yield ten of ten correct conditional-solo answers against seven with plain records, nine of ten superseded answers against eight, and one of ten retracted answers against zero. The scope result illustrates the purpose of an explicit scope field, since a preference can remain valid without applying to the current situation. Explicit records also make the remaining errors diagnosable. DeepSeek answers all other scope questions correctly but only two of ten conditional-solo questions, and the two agents answer one and two of ten retraction probes correctly although the full writer marks the revocation. These results show that correct use also depends on the agent, and that an explicit status makes a constraint available without guaranteeing that a frozen agent obeys it.

\paragraph{Interpreting the gift cell.}
Manual inspection confirms that all ten gift memories from both writers state ``a birthday gift for the user's sister.'' The checker criterion expects the gift attribution and rejects memories that present the item as the user's interest; this expectation is stricter than the history, which already contains the user's earlier self-statement. A faithful memory can therefore contain both facts. The same memories receive one of ten correct judgments from the Qwen checker and nine of ten from the DeepSeek checker. The answer key is similarly strict, marking the item-based answer wrong in the gift condition despite the earlier interest. The resulting zero of ten correct answers with Qwen and one of ten with DeepSeek diagnose how the consumer weighs gift framing against existing interest. They do not estimate pure attribution accuracy or establish that the writer misrecords the gift. We treat the automatic checks as model-dependent diagnostics cross-checked by manual inspection.

\section{Case studies}

\begin{figure}[ht]
\begin{casecard}{Case 1: Recommending buildings after a retraction}{PersonaMem-v2}
\casefield{History}{The user asked for notable buildings, and the assistant replied ``Since you love modernist architecture.'' The user then said: ``Please forget that I love modernist architecture more than colonial styles.'' The benchmark marks the query below as \code{ask\_to\_forget}.}
\casefield{Question}{``I'm planning a major renovation of my house and want to gather inspiration---could you recommend some exemplary buildings, public spaces, or design showcases that feature striking, contemporary architectural styles I could study for ideas?''}
\casefield{Options}{\textbf{A.} If you're looking for architectural inspiration, you might explore Frank Lloyd Wright's Prairie School designs for their harmony with nature, or Antoni Gaud\'i's whimsical, organic forms in Barcelona. Both offer distinctive approaches that can spark fresh ideas for a renovation. \textcolor{qabmteal}{\textbf{[key]}}\par\vspace{2pt}
\textbf{B.} Given your preference for modernist architecture over colonial styles, you could study the glass-and-steel precision of Mies van der Rohe's Farnsworth House or the sweeping, organic forms of Oscar Niemeyer's Bras\'ilia buildings.\par\vspace{2pt}
\textbf{C.} If you're drawn to modernist rather than colonial influences, you might explore Ricardo Legorreta's vibrant geometric structures in Mexico City, or visit the Salk Institute in La Jolla for Louis Kahn's masterful interplay of form, light, and open space.\par\vspace{2pt}
\textbf{D.} Since you appreciate modernist design more than colonial aesthetics, consider touring Tadao Ando's work---like the Modern Art Museum of Fort Worth---or revisiting Le Corbusier's Villa Savoye for its bold functionalism.}
\casefield{Memory}{\textit{2 of 13 records}\par\vspace{2pt}
\beliefrecord{The user has asked to forget a preference regarding Architecture; do not use or mention it}{user instruction}{always}{\statusretracted}
\beliefrecord{The user previously expressed a preference for modernist architecture over colonial styles, but has requested this preference be forgotten.}{msg 9, msg 11}{situational (architecture)}{\statusretracted}}
\casefield{Agent}{It identifies the second record as relevant and checks each option against it. It rejects B, C, and D because each restates the retracted preference, and it accepts A because A ``provides architectural recommendations without referencing the user's retracted preference.''}
\casefield{Result}{The agent outputs ``Final Answer: A'', which matches the benchmark key.}
\end{casecard}
\caption{A PersonaMem-v2 example in which the agent answers a recommendation question after the user has retracted a preference.}
\label{fig:case-answering}
\end{figure}

\label{app:cases}
\suppressfloats[t]
\makeatletter\setlength{\@fptop}{0pt}\makeatother
We select two cases, one from each benchmark, to analyze and illustrate clearly how HabitusMem works. Memory excerpts are verbatim and state how many generated records they show, and Chinese records and dialogues are translated. Together, the cases show why writing memory as beliefs helps an agent act on the user's current preferences. In Case~1, the options differ only in whether they invoke a withdrawn preference, so a correct answer requires knowing both what was withdrawn and that it was withdrawn. The retracted record supplies both. It names the preference, which lets the agent recognize three different paraphrases of it, and its status tells the agent not to use it, while leaving the agent free to answer the question about architecture. A memory that deleted the preference could not identify these paraphrases, and a memory that kept it without a status could support the wrong options. In Case~2, the request supplies none of the arguments that the purchase needs. The guidance $g_\tau$ names the constraints to extract and asks the agent to finish the transaction, but it contains no user-specific values. The records supply them, each stating one exact brand, flavor, size, or address that maps onto a tool argument, so the agent can fill the calls without asking again. The alternative record also lets the agent recover from missing stock with a choice from the user's own history. In both cases, the records state what holds about the user, where it applies, and whether it still holds, so the agent acts on explicit beliefs rather than reconstructing preferences from raw content. Stated values can also keep an action repeatable across trials.

\begin{figure}[t]
\begin{casecard}{Case 2: Buying dog food}{VitaBench 2.0}
\casefield{History}{The user wrote of her dog: ``ZIWI's food has a good nutritional profile and Bao loves it; it is expensive, but I will keep buying it until he stops liking it.'' Her stored favorites contain ``ZIWI chicken-flavored dog food, 1~kg.''}
\casefield{Request}{``I completely forgot to buy dog food for Bao---ordering from Taobao is too late; just buy it for me and deliver it home.'' The request names no brand, flavor, size, or address.}
\casefield{Guidance}{Every memory system in the comparison receives the same execution guidance $g_\tau$:\par\vspace{2pt}
{\raggedright\ttfamily\footnotesize\frenchspacing The memory above is evidence, not a response to repeat. For the current request:\par\vspace{1.5pt}
\hangindent=1.2em\hangafter=1 1. Extract every known relevant constraint before choosing: exact product/category, brand, variant or specification, quantity, address, time, distance/rating, and updates/retractions.\par\vspace{1.5pt}
\hangindent=1.2em\hangafter=1 2. Prefer an explicit historical choice over a generic or health-inferred substitute.\par\vspace{1.5pt}
\hangindent=1.2em\hangafter=1 3. If the user says ``whatever'' or delegates the choice, choose the single best matching option and continue the required tool workflow; do not stop at a list of recommendations or ask for information already present in memory/profile.\par\vspace{1.5pt}
\hangindent=1.2em\hangafter=1 4. Complete the transaction when the request asks to buy/book/order.\par}}
\casefield{Memory}{\textit{5 of 20 records}\par\vspace{2pt}
\beliefrecord{Dog food brand: ZIWI}{[msg 291], [msg 296], [msg 299]}{always}{\statusactive}
\beliefrecord{Flavor, current: chicken (currently feeding)}{[msg 296], [msg 299]}{situational (current stock)}{\statusactive}
\beliefrecord{Flavor, alternative: beef (already added to cart, pending switch)}{[msg 299]}{situational (future stock)}{\statusactive}
\beliefrecord{Format: 1~kg}{[msg 291], [msg 296], [msg 299]}{always}{\statusactive}
\beliefrecord{Delivery address: 143 Shanlin West Road, Building 8, Unit 1, Room 704}{[msg 285], [msg 288], [msg 289]}{always}{\statusactive}}
\casefield{Agent}{The search finds only ZIWI beef in 1~kg in stock, and the user replies ``Whatever, you decide; deliver it whenever.'' The tool calls draw their arguments as follows:\par\vspace{3pt}
{\footnotesize\setlength{\tabcolsep}{3pt}\renewcommand{\arraystretch}{1.05}%
\begin{tabular}{@{}>{\raggedright\arraybackslash\ttfamily\scriptsize}p{5.1cm}>{\raggedright\arraybackslash}p{3cm}>{\raggedright\arraybackslash}p{\dimexpr\linewidth-5.1cm-3cm-12pt\relax}@{}}
\toprule
\rmfamily\footnotesize\textbf{Tool call} & \textbf{Arguments} & \textbf{Source}\\
\midrule
delivery\_product\_search\_recommand & ZIWI, dog food, chicken flavor, 1~kg & Brand, current flavor, and format records\\
get\_delivery\_store\_info & The pet-food store & Search result\\
create\_delivery\_order & One bag of 1~kg ZIWI beef, full address & Alternative flavor and address records\\
pay\_delivery\_order & The created order & $g_\tau$ and user confirmation\\
\bottomrule
\end{tabular}\par}}
\casefield{Result}{The rubric checks the brand and the delivery address, and the subtask succeeds in all four trials, each paying for the same product at the same address.}
\end{casecard}
\caption{A VitaBench 2.0 example in which the agent completes a paid delivery order from an underspecified request.}
\label{fig:case-tool-use}
\end{figure}

\end{document}